\documentclass[runningheads]{llncs}
\titlerunning{}
\usepackage[T1]{fontenc}
\usepackage{graphicx}
\usepackage{amsmath,bm}
\usepackage{cite}
\usepackage{booktabs}
\usepackage{tabularx}
\usepackage{arydshln}
\title{FedDTRE: Federated Dialogue Generation Models Powered by Trustworthiness Evaluation}
\titlerunning{Model updating with Trustworthiness Evaluation}
\author{
    Shule Lu\inst{1,2} \and
    Lingxiang Wang\inst{1,2} \and
    Sijia Wen\inst{1,2} \and
    Ziwei Wang\inst{1,2} \and
    Hainan Zhang\inst{1,2} \thanks{Corresponding author: zhanghainan@buaa.edu.cn}
}
\authorrunning{S. Lu et al.}
\institute{
   Beijing Advanced Innovation Center for Future Blockchain and Privacy Computing \and 
   Institute of Artificial Intelligence, Beihang University, China \\
  \email{\{lsl2025, wanglingxiang, sijiawen, wangziwei26, zhanghainan\}@buaa.edu.cn}
}
\begin{document}
\maketitle              
\begin{abstract}
With the rapid development of artificial intelligence, dialogue systems have become a prominent form of human–computer interaction. However, traditional centralized or fully local training approaches face challenges in balancing privacy preservation and personalization due to data privacy concerns and heterogeneous device capabilities. Federated learning, as a representative distributed paradigm, offers a promising solution. However, existing methods often suffer from overfitting under limited client data and tend to forget global information after multiple training rounds, leading to poor generalization. To address these issues, we propose FedDTRE, a Federated adaptive aggregation strategy for Dialogue generation based on Trustworthiness Evaluation. Instead of directly replacing local models with the global model, FedDTRE leverages trustworthiness scores of both global and local models on a fairness-oriented evaluation dataset to dynamically regulate the global model’s contribution during local updates. Experimental results demonstrate that FedDTRE can improve dialogue model performance and enhance the quality of dialogue generation.

\end{abstract}
%
%
%
%
%
\section{Introduction}
As a core technological direction of dialogue systems, open-domain dialogue generation models have attracted extensive attention from both industry and academia~\cite{zhang2019recosa,kann-etal-2022-open}. With the emergence of large language models, dialogue generation has demonstrated remarkable improvements in semantic modeling, contextual understanding, and language generation. These advances have enabled wide-ranging applications in scenarios such as intelligent assistants~\cite{gong2023multimodal}, educational tutoring~\cite{paladines2020systematic}, medical consultation~\cite{bao2023disc}, and online customer service~\cite{yun2023fine}.


However, the construction of high-quality dialogue generation models depends on massive amounts of real user data. Although traditional centralized training can leverage large-scale data and computational resources to enhance model performance, it simultaneously raises risks of data privacy leakage and regulatory noncompliance for private scenario~\cite{mammen2021federated}. Conversely, fully localized training paradigms are constrained by the limited data volume and computational capacity of end devices, often resulting in insufficient model generalization that fails to satisfy practical application requirements. In order to reconcile privacy protection with performance optimization, Federated Learning\cite{mcmahan2017communication} (FL) has gradually emerged as an ideal training paradigm. By enabling model training on local devices while transmitting only parameters to aggregation, FL can safeguard privacy while fully exploiting distributed computational resources.

Despite its promising potential, the application of FL in dialogue generation models continues to encounter challenges~\cite{li2020federated}, particularly overfitting and the forgetting of global knowledge caused by disparities in client data scale. Dialogue data are typically stored in a distributed manner across users’ local devices. Each device contains the conversation history generated during the use of dialogue applications. As a result, while the number of user devices is large, each device only holds a relatively small amount of data. This gives rise to the characteristic of locally stored data being small-sample and dispersed. Local models trained on such small and fragmented datasets are prone to overfitting and risk losing the generalization ability of the global model~\cite{lee2022preservation,zhao2018federated}. Therefore, it is necessary to establish an appropriate mechanism for data collaboration and coordination to enhance the generalization capacity of local models.

Researchers have explored data sharing~\cite{tuor2021overcoming}, data augmentation~\cite{duan2019astraea,shin2020xor,zhu2021federated}, and model aggregation~\cite{wang2020federated,li2020federatedOP,smith2017federated,lin2020ensemble} to address non-convergence on small-sample datasets. Data sharing and augmentation approaches attempt to mitigate this issue by directly processing user data or by creating proxy data through simple transformations~\cite{duan2019astraea}, mixing~\cite{shin2020xor}, or generative adversarial networks (GANs)~\cite{zhu2021federated} for client-to-client exchange. However, such methods carry inherent risks of privacy leakage during data exchange. Model aggregation, on the other hand, employs techniques such as layer-wise aggregation~\cite{lee2023layer,wang2020federated}, regularization~\cite{acar2021federated,li2020federatedOP,smith2017federated}, and knowledge distillation~\cite{zhu2021data,lin2020ensemble} to alleviate the challenges posed by non-IID data across clients. Yet, model aggregation remains constrained by high computational overhead and limitations in model capacity. Consequently, designing federated learning models that are well-suited for extremely dispersed and heterogeneous scenarios, such as dialogue applications, remains an open and important research problem.

We find that the global model inherently contains information from peer clients’ data, which is key to enhancing the model’s generalization ability. Therefore, by introducing the global model into local training and adaptively determining its contribution through aggregation weights, the issue of weak generalization in local models can be alleviated. However, excessive reliance on the global model not only increases computational costs but may also diminish the local model’s ability to generate personalized responses. Conversely, insufficient incorporation of the global model may fail to yield a significant impact on generalization. Therefore, the timing of information integration may affect the model’s final performance.

In this paper, we propose FedDTRE, a novel Federated adaptive update strategy for Dialogue generation models, which selectively introduces global information based on Trustworthiness-based Response Evaluation. By comparing the responses generated by the local and global models, the strategy dynamically increases the aggregation weight when the global model’s response is more relevant to the dialogue context and exhibits higher trustworthiness, and decreases it otherwise. Consequently, it avoids high-cost parameter computations and instead relies only on trustworthiness evaluation of generated responses to dynamically determine the aggregation weights during local updates. 


Experimental results on Synthetic-Persona-Chat, CMU\_DoG, and WoW datasets demonstrate that FedDTRE can enhance dialogue generation quality and achieve a superior balance between privacy protection and personalized modeling. This work not only effectively alleviates the overfitting problem of small-data clients but also preserves global knowledge, thereby improving the generalization capability of the overall model. 

%
%

\section{Related Work}

Since the introduction of FedAvg~\cite{mcmahan2017communication}, Federated Learning (FL) has emerged as a pivotal framework for balancing privacy protection with distributed model training. Its paradigm of local training with global aggregation has shown promise across domains such as healthcare, finance, and dialogue systems. However, under practical conditions involving non-IID data distributions, device heterogeneity, and model heterogeneity, both performance and fairness remain constrained. To address these challenges, prior research has proposed improvements from aspects such as federated optimization~\cite{karimireddy2020scaffold,reddi2020adaptive}, and aggregation strategies~\cite{wang2020federated,singh2020model}). In parallel, Personalized FL has been proposed to enhance user-level adaptability~\cite{t2020personalized,fallah2020personalized,arivazhagan2019federated}. However, existing approaches have largely focused on general scenarios, while systematic investigations into data scarcity and global knowledge forgetting in dialogue model training remain insufficient. 

A particularly critical challenge arises when clients possess only small-scale datasets, which severely hampers convergence and model generalization. To address this issue, researchers have investigated approaches such as data sharing, data augmentation, and model aggregation. Data sharing~\cite{tuor2021overcoming} directly handles non-independent and identically distributed (non-IID) data, but it is difficult to obtain a global dataset. Data augmentation methods instead expand local training samples through trivial transformations~\cite{duan2019astraea}, mixing-based strategies~\cite{shin2020xor}, or generative adversarial networks (GANs)~\cite{zhu2021federated}. However, both of these approaches require data exchange, which introduces potential privacy leakage risks.

As a representative of model aggregation methods, FedMA \cite{wang2020federated} proposed a hierarchical aggregation strategy for non-IID partitions by sharing a global model layer by layer. The degree of aggregation is determined by calculating the consistency between each layer’s parameters and the global model, but this approach suffers from extremely high computational complexity. To address this problem, regularization-based optimization methods have been introduced to handle data heterogeneity. For instance, FedProx \cite{li2020federated} mitigates heterogeneity by incorporating proximal terms into information aggregation. MOCHA \cite{smith2017federated} takes into account communication costs, stragglers, and fault tolerance, but it is not suitable for non-convex optimization tasks. FedDF \cite{lin2020ensemble}, on the other hand, employs ensemble distillation to reduce both privacy leakage risks and computational costs in federated knowledge distillation.

Nevertheless, model aggregation approaches generally require comparing parameters between local and global models, which leads to high computational costs. Moreover, the aggregation conditions are often unrelated to the actual capacity of the models, presenting further limitations.

%
%
\begin{figure}[!t]
    \centering
    \includegraphics[width=\textwidth]{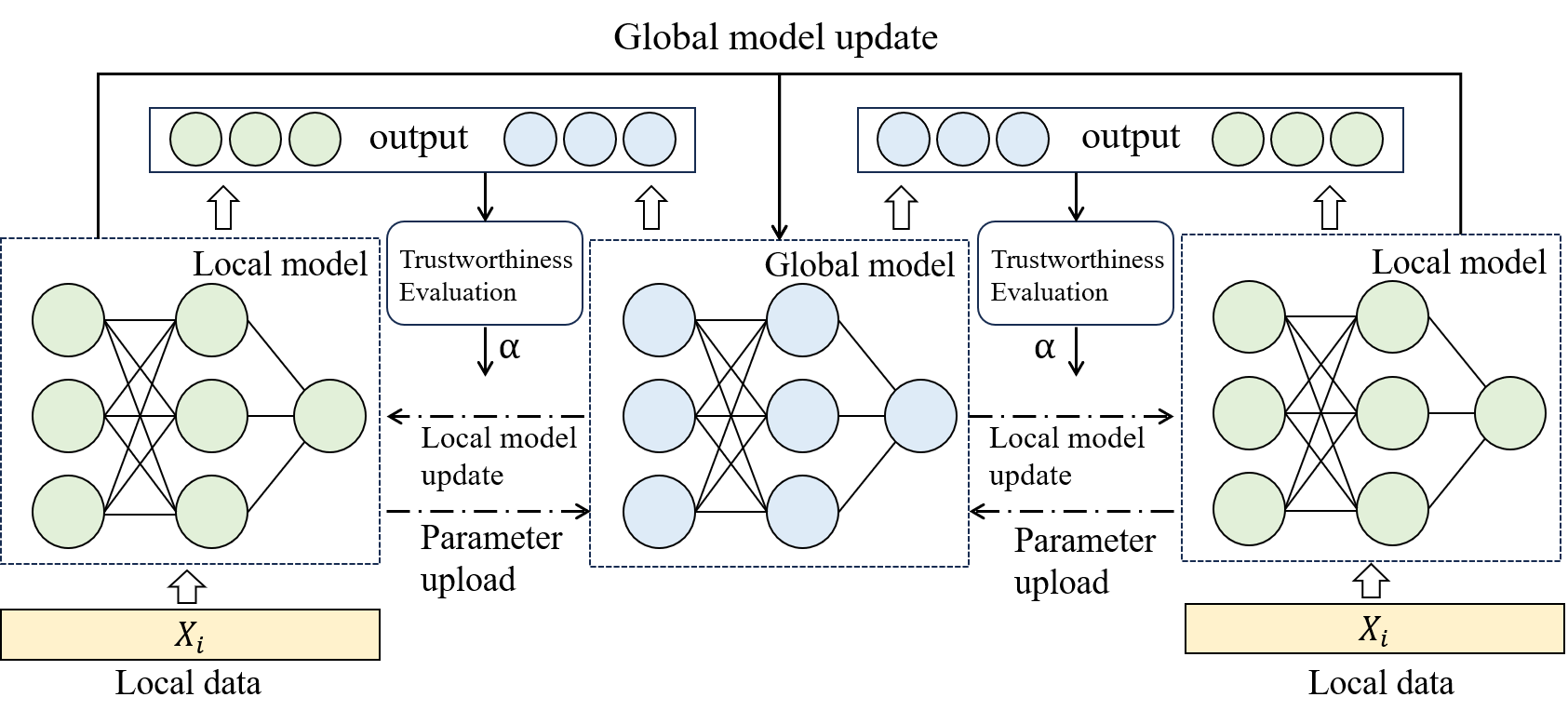}
    \caption{The overall framework of FedDTRE. It evaluates the trustworthiness of global and local models on a fairness-oriented dataset, and dynamically regulates the global model’s contribution during local updates.}
    \label{fig:framework}
\end{figure}

\section{Model}
\subsection{Trustworthiness Evaluation}
In federated dialogue learning, the evaluation of response quality relies not only on semantic relevance but also on the consideration of privacy leakage risk. To this end, we introduce Trustworthiness, which is employed to jointly assess the performance of generated responses in terms of semantic relevance and privacy compliance. A response is regarded as highly trustworthy when it can ensure strong relevance to the dialogue context while simultaneously avoiding the exposure of sensitive information.

To implement trustworthiness modeling, we build Trustworthiness Evaluation Dataset, which consists of dialogue contexts paired with candidate responses annotated with trustworthiness labels(see in Section~\ref{sec:data}). Based on this dataset, BERT model~\cite{devlin2019bert} is fine-tuned under the Federated Learning framework, thereby enabling decentralized trustworthiness modeling. Specifically, each client utilizes its local private data to train the BERT model and uploads only the parameters to the server; the server then performs Global Aggregation to obtain an optimized Global Trustworthiness Evaluator.

Formally, given a dialogue context $X$ and a candidate response $R$, we compute their contextual embeddings$\{x_{i1},\ldots,x_{ik}\}$ and $\{r_{i1},\ldots,r_{ik}\}$.Then we compute recall, accuracy, and the F1-score, and regard the F1-score of the response as the trustworthiness score, which can be defined as follows\cite{zhang2019bertscore}:

\begin{equation}
    \begin{aligned} 
    & R_{BERT}=\frac{1}{|X_i|}\sum_{x_{ip}\in X_i}\max_{r_{ij}\in R_i}x_{ip}r_{ij} \\ 
    & P_{BERT}=\frac{1}{|R_i|}\sum_{r_{ij}\in R_i}\max_{x_{ip}\in X_i}x_{ip}r_{ij} \\ 
    & F_{BERT}=2\frac{P_{BERT}\cdot R_{BERT}}{P_{BERT}+R_{BERT}} 
    \end{aligned}
\end{equation}

The trustworthiness score $S_R$ is defined as the output of the global BERT model, which can be expressed as follows: \par

\begin{equation}
    S_R = BERT_{global}(X,R)
\end{equation}

Through iterative local fine-tuning and global aggregation, the evaluator achieves robust performance while preserving user privacy. The resulting trustworthiness scores serve as an adaptive signal in the federated training stage, guiding the integration of global and local knowledge.

\begin{figure}[!t]
    \centering
    \includegraphics[width=\textwidth]{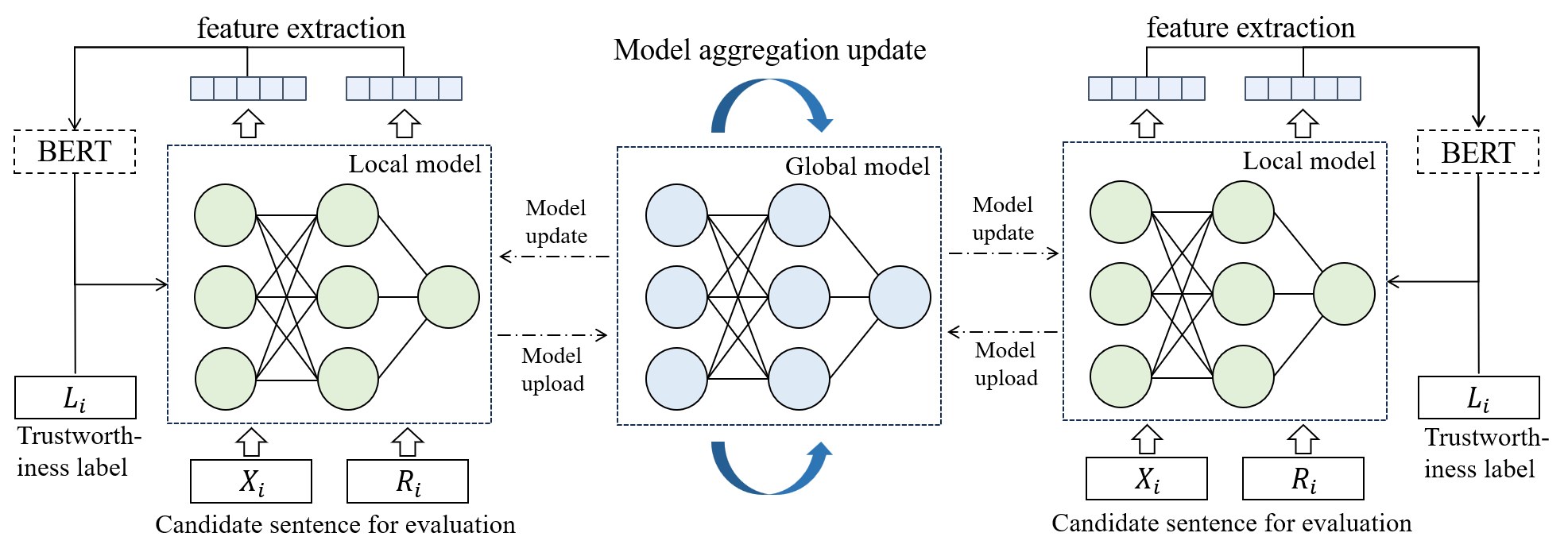}
    \caption{Federated training process of the BERT-based trustworthiness model. Each client trains the model on its local private data and uploads only model parameters to the server. The server performs global aggregation to obtain an optimized Global Trustworthiness Evaluator.}
    \label{fig:train_bertscore}
\end{figure}

\subsection{Federated Dialogue Generation}

In this method, we incorporate the evaluation of response trustworthiness into the local model updating process to adaptively adjust the participation weight of the global model during local training. This mechanism mitigates overfitting caused by limited client data and enhances the generalization capability of the overall model. The complete workflow is illustrated in Figure~\ref{fig:framework}.

During the aggregation phase, the server optimizes the global parameters $W_{\text{global}}$ with the objective of minimizing the average of the client loss functions: 

\begin{equation}
    F(W_{global}) = \frac{1}{K}\sum_{i\in S_t}f_i(W_{local}^i,D_i)
\end{equation}

\noindent where $D_i$ is the local dataset on the client, and $f_i$ is the local loss function. $S_t\subseteq\{1,2,\dots,K\}$ denotes the active client subset participating in training round $t$ and $K$ equals the cardinality of that subset. During model updates, each client combines the global parameters with the locally trained results: 

\begin{equation}
    W_{local}^{i^*}=(1-\alpha)\underset{D_i}{\operatorname*{\operatorname*{min}}}f(W_{\mathrm{local}}^i)+\alpha W_{\mathrm{global}}
\end{equation}

\noindent where the weighting factor $\alpha$ is determined by the score difference  between the local model $M_{local}^i$ and the global model $ M _ {global} $ on the Trustworthiness Evaluation Dataset, which can be defined as follows: 

\begin{equation}
    \alpha = \alpha_{min} + \varphi_{score}^{i} \cdot (\alpha_{max}-\alpha_{min})
\end{equation}

\noindent where $\varphi_{score}^i $ is calculated based on the sigmoid function:

\begin{equation}
    \varphi_{\mathrm{score}}^i=
    \begin{cases}
        0, & \mathrm{if~\Delta s_i}\leq0 \\
        \frac{\sigma(\Delta s_i)-\sigma(0)}{\sigma(1)-\sigma(0)}, & \mathrm{if~\Delta s_i}>0 
    \end{cases}
\end{equation}

\noindent where $\sigma(x)$ and $\Delta s_i$ can be defined as follows:

\begin{equation}
    \sigma(x) = \frac{1}{1 + e^{-k(x-midpoint)}}
\end{equation}
\begin{equation}
    \Delta s_i = s_g - s_l^i
\end{equation}

 \noindent where $k$ controls the steepness of the sigmoid function, and it increases as the number of training epochs grows, $s_g$ is the score of the global model $M_{global}$ on the Trustworthiness Evaluation Dataset, and $s_l^i$ is the score of the local model:

\begin{equation}
    \begin{aligned}
        & s_l^i = BERT_{global}^*(X,M_{local}^i(X)) \\
        & s_g = BERT_{global}^*(X,M_{global}(X))
    \end{aligned}
\end{equation}

After the local updates are completed, the server performs global aggregation using a strategy similar to FedAvg:

\begin{equation}
    W_{global}^{i^*} = \frac{1}{K}\sum_{i \in S_t} W_{local}^{i^*}
\end{equation}

Finally, this process is repeated over multiple iterations until the model converges or reaches the predetermined number of training rounds.

%
%
\section{Experiments}
\subsection{Experimental Settings}
In this section, we described the training and model evaluation settings of federated learning.
\subsubsection{Evaluation Datasets}
This study involves four types of dialogue datasets, serving federated learning, trustworthiness evaluation, and knowledge-grounded dialogue tasks.

For the FL client scenarios, we use the Synthetic-Persona-Chat dataset\cite{jandaghi2023faithful}, which is an extension of Persona-Chat\cite{zhang2018personalizing}. It contains 4,723 original personas with 10,906 dialogues and 5,648 synthetic personas with 11,001 dialogues, overall exhibiting persona-driven characteristics.

In addition, we introduce two representative knowledge-grounded dialogue datasets for evaluation. The first is CMU\_DoG\cite{zhou2018dataset}, constructed based on Wikip-edia documents, which emphasizes document-grounded conversations, comprising approximately 4,000 dialogues and highlighting factual accuracy and context dependency. The second is Wizard of Wikipedia (WoW)\cite{dinan2018wizard}, released by FAIR, containing around 22,000 dialogues and adopting a Wizard–Apprentice setting that integrates persona information with external knowledge, enabling the model to generate more informative and personalized responses.

\subsubsection{Trustworthiness Dataset}\label{sec:data}

To train a BERT model capable of evaluating the trustworthiness score, we build a Trustworthiness Dialogue Dataset. This dataset is derived from pii-masking-300k\cite{ai4privacy_2024} , which integrates OpenPII-220k--which contains 27 types of PII and approximately 220,000 instances spanning the education, healthcare, and mental health domains--and FinPII-80k, comprising over 20 types of financial and insurance-related PII with approximately 80,000 instances. 

The dataset construction proceeds as follows: First, we concatenate the source text with privacy masks and input them into a general LLM to generate queries targeting sensitive entities. Second, conditioned on these queries, the LLM generates positive and negative trustworthiness pairs. Finally, we score the candidate responses for trustworthiness using DeepSeek-r1 to form a fully annotated dataset. This dataset provides high-quality supervision signals for subsequent federated fine-tuning.

\subsubsection{Training Details}

We use DeepSeek-LLM-7B-Chat~\cite{bi2024deepseek} as the base dialogue model, quantized to 4 bits~\cite{banner2019post}. DeepSeek-LLM-7B-Chat is a large language model-based dialogue model, specifically, it is a fine-tuned model with 7B parameters, optimized for dialogue generation tasks. We train and fine-tune the model using the PyTorch deep learning framework together with the widely adopted LoRA fine-tuning strategy~\cite{hu2022lora}.

This study employs the Flower federated learning framework. During local client training, we set the training step to 5, configure QLoRA with $lora_r = 8$ and $lora_\alpha=16$, and use a batch size of 16 with a maximum sequence length of 512. We set the local learning rate to range from a maximum of $1\times10^{-61}$  to a minimum of $5\times10^{-55}$. Each client performs training on a single A100 GPU. On the server side, we run 100 communication rounds, randomly selecting two clients for local training in each round. For trustworthiness evaluation between the client and server models, we sample 100 instances from the Trustworthiness Evaluation Dataset per round. In the computation of $\alpha$, we set the initial $k$ to 0.7, the midpoint to 0.01, the $\alpha_{max}$ to 1, and the $\alpha_{min}$ to 0.1.

\subsubsection{Evaluation Metrics}

To quantitatively evaluate the effectiveness of dialogue text generation, we employ three evaluation metrics: BLEU , ROUGE , and the BERTScore. BLEU\cite{papineni2002bleu} and ROUGE\cite{lin2004rouge} measure the semantic relevance between the generated responses and the reference texts, whereas the BERTScore is used to evaluate the trustworthiness of the responses. 
BLEU\cite{papineni2002bleu} is a widely used automatic evaluation metric for machine translation and text generation. Its core idea is to compare the n-gram overlap between the generated text and the reference text, with BLEU-1 to BLEU-4 representing matches at different granularities. To prevent the model from exploiting excessively short outputs, BLEU introduces a brevity penalty (BP). The final score ranges from 0 to 1, with higher scores indicating that the generated text is closer to the reference text.

\begin{equation}
    \begin{aligned}
        BLEU & = BP\times\exp\left(\sum_{n=1}^NW_n\times\log P_n\right) \\
        BP  & =
        \begin{cases}
            1 & lc>\mathrm{lr} \\
            \exp{(1-lr/lc)} & lc\leq lr 
        \end{cases}
    \end{aligned}
\end{equation}

\noindent where BP denotes the brevity penalty, and $P_n$ represents the n-gram precision score. 

ROUGE-N\cite{lin2004rouge} is the most commonly applied form within the ROUGE family of metrics and is primarily used to assess the quality of automatic text summarization and natural language generation tasks. This metric calculates the n-gram overlap rate between the generated text and the reference text to capture their degree of similarity. Its focus lies in measuring recall, namely, the proportion of information in the reference text that is covered by the generated text. Specifically, ROUGE-1 emphasizes word-level matching, while ROUGE-2 focuses on consecutive bigram matching. Higher ROUGE-N scores indicate that the generated text preserves more of the critical information contained in the reference text, thereby reflecting superior generation quality.

\begin{equation}
    ROUGE\mathrm-N=
        \frac{\sum_{S\in\{\text{Reference Summaries}\}}\sum_{\mathrm{gram}_n\in S}\mathrm{Count}_{match}(\mathrm{gram}_n)}
        {\sum_{S\in\{\text{Reference Summaries}\}}\sum_{\mathrm{gram}_n\in S}\mathrm{Count}(\mathrm{gram}_n)}
\end{equation}

In addition, we use the BERTScore to evaluate the trustworthiness of generated responses. Specifically, we apply the BERT model trained on the Trustworthiness Evaluation Dataset to compute the trustworthiness score of the generated text.

\begin{table}[!t]
\caption{Evaluation results on three dialogue datasets.}
\centering
\renewcommand{\arraystretch}{1.2}
\setlength{\tabcolsep}{4pt}
\begin{tabularx}{\linewidth}{lccccc}
    \toprule
    Method & BLEU-1 & BLEU-4 & ROUGE-1 & ROUGE-2 & BERTScore \\
    \midrule
    \multicolumn{6}{c}{\textit{\small Wizard of Wikipedia (WoW)}} \\[2pt]
    DeepSeek-LLM-7B-Chat & 4.53 & 0.51 & 6.40 & 1.22 & 78.70 \\
    FedAvg               & 4.34 & 0.58 & 6.13 & 1.09 & 78.54 \\
    FedProx              & 4.29 & 0.55 & 6.04 & 1.05 & 78.48 \\
    FedDTRE              & 4.00 & 0.49 & 5.66 & 0.88 & 78.57 \\
    \hdashline
    \multicolumn{6}{c}{\textit{\small CMU\_DoG}} \\[2pt]
    DeepSeek-LLM-7B-Chat & 6.82 & 0.77 & 9.25 & 1.31 & 63.89 \\
    FedAvg               & 6.76 & 0.76 & 9.07 & 1.22 & 63.81 \\
    FedProx              & 6.76 & 0.75 & 9.07 & 1.27 & 63.73 \\
    FedDTRE              & 6.85 & 0.76 & 9.04 & 1.21 & 63.77 \\
    \hdashline
    \multicolumn{6}{c}{\textit{\small Synthetic-Persona-Chat}} \\[2pt]
    DeepSeek-LLM-7B-Chat & 13.07 & 3.89 & 16.54 & 6.89 & 52.92 \\
    FedAvg               & 12.45 & 3.61 & 15.95 & 6.54 & 53.57 \\
    FedProx              & 12.31 & 3.35 & 16.10 & 6.48 & 53.83 \\
    FedDTRE              & 13.56 & 4.07 & 17.42 & 7.41 & 53.05 \\
    \bottomrule
\end{tabularx}
\label{tab:all_results}
\end{table}

\subsubsection{Baselines}

In this work, we select two federated learning algorithms for comparison, which are introduced as follows:

FedAvg\cite{mcmahan2017communication} is a classical algorithm in federated learning. In each training round, the server distributes the training configuration and model parameters, and the clients train the model locally after receiving the parameters. The clients then upload the trained model parameters back to the server. The server aggregates the parameters collected from all clients and redistributes the updated global model to them. This process is repeated until the model converges or the predefined number of communication rounds is reached.

FedProx\cite{fedprox2020federated} is an improved federated learning optimization algorithm that extends the classical FedAvg. Its improvement lies in adding a “proximal term” to the local loss function of each client, which acts as a regularization term. This mechanism constrains the deviation between the local models and the global model, thereby enhancing the stability of convergence.

\subsection{Main Results}
As shown in Table~\ref{tab:all_results}, FedDTRE demonstrates competitive performance across the three datasets. On the Synthetic-Persona-Chat dataset, FedDTRE achieves the best results on all BLEU and ROUGE sub-metrics, surpassing FedAvg and FedProx, which highlights its effectiveness in enhancing lexical-level generation quality and capturing key contextual information. Similar improvements are observed on the CMU\_DoG dataset, where FedDTRE achieves strong results, particularly in ROUGE metrics, indicating its capability to extract and organize salient information more accurately in multi-turn conversational contexts. These findings confirm that FedDTRE is able to improve dialogue relevance and maintain information consistency across turns.

However, on the Wizard of Wikipedia (WoW) dataset, FedDTRE performs less favorably compared to the baselines. A plausible explanation is that WoW is heavily knowledge-grounded and relies on precise factual alignment with external knowledge sources. Since FedDTRE emphasizes trustworthiness-aware updates, the model may focus more on semantic reliability rather than factual surface-level alignment, leading to weaker lexical overlap (as reflected in BLEU and ROUGE) despite comparable semantic adequacy.

Meanwhile, the BERTScore results across datasets show only marginal differences among methods, suggesting that all approaches achieve a similar level of semantic fidelity. Nevertheless, FedDTRE still maintains advantages on relevance-oriented metrics, implying that while BERTScore emphasizes semantic similarity, it may overlook fine-grained lexical and structural improvements brought by FedDTRE. Interestingly, on WoW, FedDTRE attains the highest BERTScore, which could be attributed to the dataset’s incorporation of user-specific knowledge. This introduces privacy-sensitive contextual features that align well with FedDTRE’s trustworthiness-oriented training mechanism, thereby enhancing semantic reliability.

Compared with the original DeepSeek-LLM-7B-Chat model, both FedAvg and FedProx often show degradation on multiple metrics across datasets, in some cases even performing worse than the non-fine-tuned baseline. This drop may result from overfitting to limited client data, which reduces the generalization ability of the global model. By contrast, FedDTRE effectively mitigates this issue through its trustworthiness evaluation mechanism, which regularizes local updates and improves the robustness of federated optimization.

\subsection{Ablation Study}

In the ablation study, we fixed the fusion coefficient $\alpha$ instead of adjusting it dynamically. As shown in Table~\ref{tab:Ablation_Study}, fixed $\alpha$ generally leads to lower or less consistent performance across datasets. On Wizard of Wikipedia, both $\alpha=0.5$ and $\alpha=0.25$ reduce BLEU and ROUGE scores compared to dynamic fusion, indicating that static fusion limits the model’s ability to balance trustworthiness and lexical accuracy. On CMU\_DoG, a smaller $\alpha$ slightly improves BLEU and ROUGE, but differences are minor. For Synthetic-Persona-Chat, $\alpha=0.5$ marginally outperforms $\alpha=0.25$. These results suggest that dynamic adjustment of $\alpha$ based on trustworthiness enables more consistent improvements in dialogue relevance and information retention than any fixed setting.

We conduct an ablation experiment in which we disable trustworthiness evaluation and set the model fusion coefficient $\alpha$ to a fixed value for federated learning training and fine-tuning. The results of the ablation experiment are presented in Table \ref{tab:Ablation_Study}.


%
%

%
%

\begin{table}[!t]
\caption{Results of the ablation study where the trustworthiness evaluation module is disabled and the model fusion coefficient $\alpha$ is fixed to 0.5 and 0.25 during federated learning training and fine-tuning. The resulting models are evaluated on three dialogue datasets.}
\centering
\renewcommand{\arraystretch}{1.2}
\setlength{\tabcolsep}{4pt}
\begin{tabularx}{\linewidth}{l*{5}{>{\centering\arraybackslash}X}}
    \toprule
    Fixed $\alpha$ & BLEU-1 & BLEU-4 & ROUGE-1 & ROUGE-2 & BERTScore \\
    \midrule
    \multicolumn{6}{c}{\textit{\small Wizard of Wikipedia (WoW)}} \\[2pt]
    $\alpha = 0.5$  & 3.89 & 0.48 & 5.38 & 0.86 & 78.52  \\
    $\alpha = 0.25$ & 3.59 & 0.47 & 5.17 & 0.79 & 78.54 \\
    \hdashline
    \multicolumn{6}{c}{\textit{\small CMU\_DoG}} \\[2pt]
    $\alpha = 0.5$  & 6.82 & 0.81 & 9.29 & 1.39 & 63.98  \\
    $\alpha = 0.25$ & 7.01 & 0.84 & 9.57 & 1.49 & 63.88 \\
    \hdashline
    \multicolumn{6}{c}{\textit{\small Synthetic-Persona-Chat}} \\[2pt]
    $\alpha = 0.5$  & 13.14 & 3.78 & 16.84 & 7.00 & 52.07  \\
    $\alpha = 0.25$ & 13.02 & 3.73 & 16.68 & 6.84 & 51.70 \\
    \bottomrule
\end{tabularx}
\label{tab:Ablation_Study}
\end{table}


\section{Conclusion}
We proposed FedDTRE, a federated adaptive update strategy for dialogue generation models that leverages Trustworthiness-based Response Evaluation to balance global and local model contributions. By dynamically adjusting aggregation weights based on response quality, FedDTRE improves generalization without heavy computation or privacy risks. Experiments on Synthetic-Persona-Chat, CMU\_DoG, and WoW datasets show that our method alleviates overfitting on small-data clients, preserves global knowledge, and enhances dialogue quality. FedDTRE thus offers an effective and lightweight solution for federated dialogue generation in heterogeneous and privacy-sensitive settings.
Future work will explore extending FedDTRE to multimodal and cross-lingual dialogue systems while incorporating more advanced trustworthiness metrics for broader real-world applicability.

\bibliographystyle{splncs04}

\bibliography{ref} 

%




\end{document}